\documentclass{article}

\usepackage{arxiv}

\usepackage[utf8]{inputenc} 
\usepackage[T1]{fontenc}    
\usepackage{hyperref}       
\usepackage{url}            
\usepackage{booktabs}       
\usepackage{amsfonts}       
\usepackage{nicefrac}       

\usepackage{orcidlink}
\usepackage{cite}
\usepackage{amsmath,amssymb,amsfonts}
\usepackage{multirow} 
\usepackage{algpseudocode}
\usepackage{graphicx}
\usepackage{textcomp}
\usepackage{xcolor}
\usepackage[linesnumbered,ruled,vlined]{algorithm2e}
\usepackage{amsthm}

\usepackage{threeparttable}
\newtheorem{theorem}{Theorem}
\newtheorem{definition}{Definition}
\newtheorem{prop}{Proposition}
\usepackage[normalem]{ulem}

\newtheorem{Rem}{Remark}
\usepackage{microtype}      
\usepackage{lipsum}
\usepackage{graphicx}
\graphicspath{ {./images/} }

\title{Differentially Private Decentralized Dataset Synthesis Through Randomized Mixing with Correlated Noise}

\author{
 Utsab Saha\orcidlink{0000-0003-2106-8648}\\
  School of Data and Sciences\\
  BRAC University\\
  Dhaka 1212, Bangladesh \\
  \texttt{utsab.saha@bracu.ac.bd} \\
   \And
 Tanvir Muntakim Tonoy\orcidlink{0009-0005-9915-5014} \\
  Electrical and Computer Engineering\\
  University of California, Santa Barbara\\
  Santa Barbara, CA 93106, USA \\
  \texttt{tanvirmuntakim@ucsb.edu} \\
  \And
 Hafiz Imtiaz\orcidlink{0000-0002-2042-5941} \\
  Electrical and Electronic Engineering\\
  Bangladesh University Engineering and Technology \\
  Dhaka 1205, Bangladesh \\
  \texttt{hafizimtiaz@eee.buet.ac.bd} \\
}

\begin{document}
\maketitle
\begin{abstract}
In this work, we explore differentially private synthetic data generation in a decentralized-data setting by building on the recently proposed Differentially Private Class-Centric Data Aggregation (DP-CDA). DP-CDA synthesizes data in a centralized setting by mixing multiple randomly-selected samples from the same class and injecting carefully calibrated Gaussian noise, ensuring ($\epsilon$, $\delta$)-differential privacy. When deployed in a decentralized or federated setting, where each client holds only a small partition of the data, DP-CDA faces new challenges. The limited sample size per client increases the sensitivity of local computations, requiring higher noise injection to maintain the differential privacy guarantee. This, in turn, leads to a noticeable degradation in the utility compared to the centralized setting. To mitigate this issue, we integrate the Correlation-Assisted Private Estimation (CAPE) protocol into the federated DP-CDA framework and propose CAPE Assisted Federated DP-CDA algorithm. CAPE enables limited collaboration among the clients by allowing them to generate jointly distributed (anti-correlated) noise that cancels out in aggregate, while preserving privacy at the individual level. This technique significantly improves the privacy-utility trade-off in the federated setting. Extensive experiments on MNIST and FashionMNIST datasets demonstrate that the proposed CAPE Assisted Federated DP-CDA approach can achieve utility comparable to its centralized counterpart under some parameter regime, while maintaining rigorous differential privacy guarantees.
\end{abstract}

\footnote{This work has been submitted to the IEEE for possible publication. Copyright may be transferred without notice.}
\keywords{Differential privacy \and decentralized computation \and Data synthesis \and correlated noise \and federated learning algorithm}

\section{Introduction}
As machine learning (ML) systems expand into sensitive domains, such as healthcare, finance, and mobile applications, data privacy concerns have intensified. Additionally, sensitive data are often stored in a decentralized manner on multiple nodes. Pooling such data in a centralized location poses significant privacy risks. Several studies have shown that trained models can (un)intentionally leak private information through various attacks, including membership inference, model inversion, and property inference~\cite{shokri2017membership, nasr2019comprehensive, carlini2019secret, ganju2018property}. Federated Learning (FL) has emerged as a promising solution to address these concerns, as a privacy-promoting and communication-efficient distributed alternative to centralized learning~\cite{chen2023federated}. In FL, multiple clients collaboratively train a shared global model while keeping their data local~\cite{augenstein2019generative, li2020federated, yang2019federated}. Only model updates, such as gradients or weights, are exchanged with a central server, which reduces the risk of raw data exposure~\cite{mcmahan2017communication, konevcny2016federated}. This setup makes FL particularly attractive for privacy-sensitive applications. However, FL is not immune to privacy-threats. Despite not sharing raw data, model updates can still reveal sensitive information if not properly protected. Even more concerning are attacks specific to decentralized systems, where adversaries exploit intermediate model updates shared during training to reconstruct or extract sensitive data~\cite{geiping2020inverting, aouedi2022handling,jin2021cafe}.
\begin{figure}[t]
    \centering
    \includegraphics[width=1.0\textwidth]{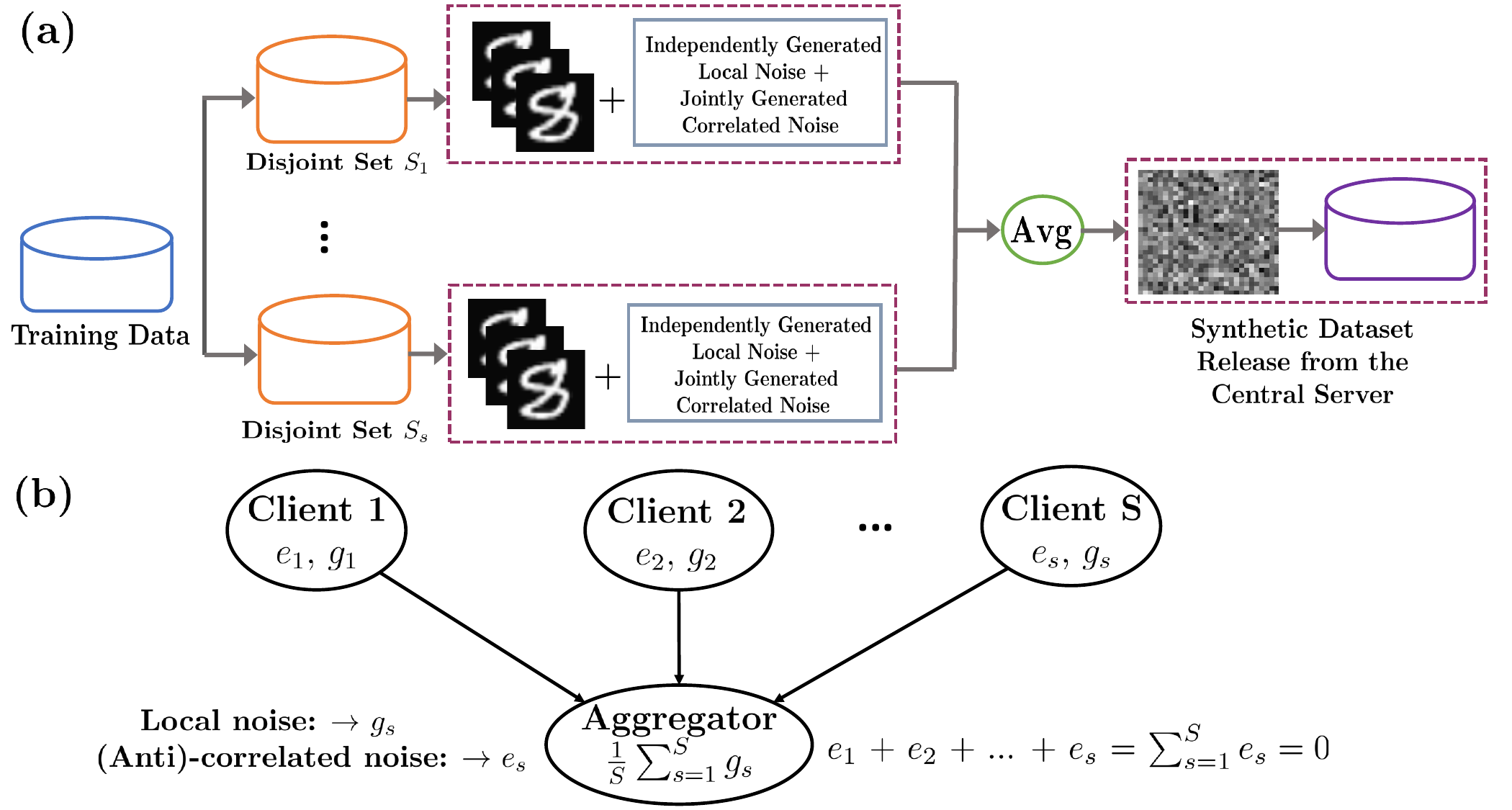}
    \caption{Correlation-assisted private estimation (CAPE) based decentralized dataset synthesis through randomized mixing (a) Data generation (b) Correlated noise cancellation process}
    \label{fig:fig2}
    \vspace{-0.7cm}
\end{figure}
To strengthen privacy in FL, different privacy-preserving FL methods have been widely adopted~\cite{xu2024tapfed, chang2023privacy, asoodeh2021differentially, xu2021fedv}. Among these privacy-preserving FL methods, differential privacy (DP) based FL methods ensure that model updates are obfuscated with calibrated noise before transmission, providing formal privacy guarantees~\cite{geyer2017differentially, asoodeh2021differentially}. While effective in theory, these approaches often suffer from unacceptable privacy-utility trade-off, i.e., achieving strong privacy requires injecting high levels of noise, which degrades model accuracy~\cite{geiping2020inverting}.

To overcome these limitations, researchers have turned to privacy-preserving data synthesis as an alternative paradigm. Instead of perturbing updates, this approach generates synthetic datasets that preserve the statistical structure of real data, but do not contain any actual private information. If the synthetic data is generated under DP constraints, it inherits the post-processing invariance property of differential privacy. That is, the synthetic data can be reused in multiple downstream tasks without consuming additional privacy budget. Moreover, synthetic data generation decouples privacy protection from the model training process, making it easier to deploy standard ML pipelines without redesigning algorithms. As privacy regulations tighten and adversarial threats grow more sophisticated, approaches that embed privacy into the data itself -- rather than the training mechanism-- are becoming increasingly vital for practical, scalable, and trustworthy ML.

Motivated by this, we intend to integrate differentially private synthetic data generation into the federated learning pipeline -- leveraging a recently proposed class-wise mixing algorithm Differentially Private Class-Centric Data Aggregation (DP-CDA)~\cite{saha2024dp} with a correlated noise injection protocol~\cite{imtiaz2021correlated} to ensure superior privacy-utility trade-off. Unlike traditional differentially private federated learning methods that inject noise into gradients and train models, our approach generates synthetic data samples that are differentially private by design. Each client randomly selects $l$ samples per class, mixes them locally, perturbs them with carefully tuned Gaussian noise, and transmits the resulting synthetic data to a central server, which aggregates the synthetic samples received from all clients. We observe that this integration of DP-CDA in a federated setting leads to a drop in utility compared to centralized setups, due to the increased noise required when each client has access to only a small portion of the total dataset. To mitigate this, we incorporate the Correlation-Assisted Private Estimation (CAPE) protocol~\cite{imtiaz2021correlated}, which allows clients to sample noise from an (anti) correlated distribution. This structured noise cancels out during aggregation, improving the privacy-utility trade-off. Our results show that even with limited data per client, the proposed CAPE Assisted Federated DP-CDA achieves utility comparable to the centralized version, highlighting its effectiveness and scalability for private, data-scarce federated environments.


\noindent\textbf{Notation. }We use lower-case bold-faced letters (e.g., $\mathbf{x}$) for vectors, upper-case bold-faced letters (e.g., $\mathbf{X}$) for matrices, and unbolded letters (e.g., $N$ or $n$) for scalars. The set of $N$ data points $(\mathbf{x}_i, y_i)$ is represented as $\mathbb{D} = \{(\mathbf{x}_1, y_1), (\mathbf{x}_2, y_2), \ldots, (\mathbf{x}_N, y_N)\} \triangleq \{ \mathbf{x}_i, y_i \}_{i=1}^N$. We use $\| \cdot \|_2$ and $\| \cdot \|_F$ to denote the $\ell_2$-norm of a vector and the Frobenius norm of a matrix, respectively.

\section{Background and Problem Formulation}\label{sec:Definitions}
We review the essential definitions, theorems, and propositions~\cite{imtiaz2021correlated, saha2024privacy, tasnim2023approximating} for our problem formulation in Appendix \ref{sec:appenA}.

\noindent\textbf{Problem Formulation.} As mentioned before, modern ML systems increasingly rely on decentralized privacy-sensitive data. To that end, FL has emerged as a promising approach to training models without sharing raw data. However, even in FL, the exchange of model parameters can still lead to significant privacy leakage~\cite{geiping2020inverting, aouedi2022handling,jin2021cafe}. Motivated by this, we focus on generating \emph{differentially private synthetic data} in a federated setting. More specifically, our goal is to create a synthetic dataset that (i) satisfies strict \((\epsilon, \delta)\)-DP guarantees across all clients, and (ii) enables superior privacy-utility trade-off in downstream machine learning tasks, comparable to that of centralized training for some parameter choices.

Suppose we have a global dataset \(\mathbb{D} = \bigcup_{s=1}^{S} \mathbb{D}_s\), where each client \(s \in \{1, \dots, S\}\) holds a private local partition \(\mathbb{D}_s = \{ (\mathbf{x}_{s,i}, \mathbf{y}_{s,i}) \}_{i=1}^{N_s}\), with \(\mathbf{x}_{s,i} \in \mathbb{R}^{d_x}\) and \(\mathbf{y}_{s,i} \in \mathbb{R}^{d_y}\). The objective is to collaboratively construct a \textbf{global synthetic dataset} \(\tilde{\mathbb{D}} = \{ (\bar{\mathbf{x}}_t, \bar{\mathbf{y}}_t) \}_{t=1}^T\), which satisfies DP and enables effective training of an ML model \(f(\mathbf{w})\). To ensure practical utility, we require that the performance of the model trained on \(\tilde{\mathbb{D}}\) remains within a predefined threshold \(\theta \in (0,1]\) compared to the performance on the original (non-private) dataset:
\begin{align}\label{eqn:utility}
\text{Utility}(f, \tilde{\mathbb{D}}) \geq \theta \cdot \text{Utility}(f, \mathbb{D}).    
\end{align}
We note that, since each client typically hosts a smaller number of data samples (compared to the aggregate number of data points in a hypothetical centralized scenario), the required noise to ensure the same level of privacy as the centralized setting is much higher~\cite{imtiaz2021correlated}. To address the degradation in utility due to this data fragmentation and higher noise requirements in the federated setting, we propose to incorporate the CAPE protocol~\cite{imtiaz2021correlated}. CAPE enables the clients to sample jointly distributed (anti-)correlated noise such that i) local privacy is preserved, and ii) the aggregate noise cancels out at the server, thereby significantly improving the utility of the global synthetic dataset $\tilde{\mathbb{D}}$ without weakening individual privacy guarantees. More specifically, in our federated DP-CDA framework, each client \(s\) mixes \(l\) samples per class from its local dataset, followed by the addition of noise according to the CAPE protocol to ensure local differential privacy and global utility. The overall design problem now becomes finding a transformation mechanism \(\mathcal{M} = \{ \mathcal{M}_1, \dots, \mathcal{M}_S \}\), where each \(\mathcal{M}_s\) operates on client \(s\)'s local data and satisfies differential privacy, while satisfying \eqref{eqn:utility}. Formally, we aim to:
\[
\text{minimize} \quad \epsilon \quad \text{(overall privacy budget)}
\]
\[
\text{subject to} \quad \text{Utility}(f, \tilde{\mathbb{D}}) \geq \theta \cdot \text{Utility}(f, \mathbb{D})
\]
By combining locally generated synthetic data with correlated noise through CAPE, our proposed framework allows maintaining strict privacy while significantly narrowing the utility gap between federated and centralized data generation.

\section{Proposed Decentralized Synthetic Data Generation: CAPE Assisted Federated DP-CDA}
In this section, we describe the proposed CAPE assisted federated DP-CDA algorithm in detail. We note that the proposed algorithm is based on the recently proposed DP-CDA~\cite{saha2024dp} algorithm and the CAPE framework~\cite{imtiaz2021correlated}. As mentioned before, our motivation is to extend the synthetic data generation process to a decentralized-data setting (or federated setting), where the sensitive dataset is partitioned across $S$ different nodes (e.g., hospitals, organizations, or clients). Each node independently performs privacy-preserving data mixing and sends noisy synthetic data to a central aggregator. The aggregator then computes the average of these data entries to generate the final synthetic dataset. All parties are honest but curious -- that is, they follow the protocol honestly, but may collude with external/internal adversaries to learn about other client's data. 
We assume each client independently performs local data preprocessing, synthetic data generation, and transmits its differentially private synthetic samples to a central aggregator.

\noindent\textbf{Local Data Preprocessing. }
Let the global dataset be denoted as \( \mathbb{D} = \{(\mathbf{x}_i, y_i)\}_{i=1}^{N} \), where \( \mathbf{x}_i \in \mathbb{R}^{d_x} \) and \( y_i \in \{1, 2, \dots, K\} \) for \( K \) classes. This dataset is partitioned across \( S \) clients, such that client \( s \in \{1, \dots, S\} \) holds a disjoint local dataset \( \mathbb{D}_s = \{(\mathbf{x}_{s,i}, y_{s,i})\}_{i=1}^{N_s} \) of size $N_s$, and  \(\sum_{s=1}^{S} N_s = N\). For simplicity, we assume $N_s = N/S$~\cite{imtiaz2021correlated}. Each client aims to generate \( T_s \) synthetic samples. Without loss of generality, we assume that the aggregator intends to release a synthetic dataset of size \( T = T_s \). Each client begins by applying the following preprocessing steps to their local data:
\begin{enumerate}
    \item \textbf{Z-score normalization} on each feature dimension as $x_{s,ij} \gets \dfrac{x_{s,ij} - \mu_{s,j}}{\sigma_{s,j}}$ $\forall j \in \{1, \dots, d_x\}$, where \( \mu_{s,j} \) and \( \sigma_{s,j} \) are the local mean and standard deviation computed at client \( s \).
        
    \item \textbf{\(\ell_2\) norm clipping} as $\mathbf{x}_{s,i} \gets \dfrac{\mathbf{x}_{s,i}}{\max\left(1, \|\mathbf{x}_{s,i}\|_2 / c\right)}$, where \( c \) is a predefined clipping threshold.
\end{enumerate}
Let \( \mathbf{X}_s \in \mathbb{R}^{n_s \times d_x} \) and \( \mathbf{Y}_s \in \mathbb{R}^{n_s \times K} \) denote the locally preprocessed feature matrix and one-hot encoded label matrix at client \( s \), respectively.

\noindent\textbf{CAPE-assisted Synthetic Data Generation. }Now, for each class \( k \in \{1, \dots, K\} \), the client generates the $t$-th synthetic sample for $t\in\{1, 2, \ldots, T_k\}$, where \( T_k = T_s / K \), by randomly selecting \( l \) samples from the local class-specific set $\mathbf{X}_{s,k} = \{\mathbf{x}_{s,i} \in \mathbf{X}_s : y_{s,i} = k\}$ as: $\tilde{\mathbf{x}}^{(s,k)}_t = \frac{1}{l} \sum_{j=1}^{l} \mathbf{x}_{s,i_j} + \mathbf{e}_{s,x}^{(t)} + \mathbf{g}_{s,x}^{(t)}$ and $\tilde{\mathbf{y}}^{\text{one-hot}(s,k)}_t = \frac{1}{l} \sum_{j=1}^{l} \mathbf{y}^{\text{one-hot}}_{s,i_j} + \mathbf{e}_{s,y}^{(t)} + \mathbf{g}_{s,y}^{(t)}$, where:
\begin{itemize}
    \item \( \mathbf{g}_{s,x}^{(t)} \sim \mathcal{N}(0, \tau_g^2 \mathbf{I}_{d_x}) \), and \( \mathbf{g}_{s,y}^{(t)} \sim \mathcal{N}(0, \tau_g^2 \mathbf{I}_K) \) are independently generated local noise vectors.
    \item \( \mathbf{e}_{s,x}^{(t)} \sim \mathcal{N}(0, \tau_e^2 \mathbf{I}_{d_x}) \), and \( \mathbf{e}_{s,y}^{(t)} \sim \mathcal{N}(0, \tau_e^2 \mathbf{I}_K) \) are correlated noise vectors jointly generated across clients such that $\sum_{s=1}^S \mathbf{e}_{s,x}^{(t)} = \mathbf{0}$ and $\sum_{s=1}^S \mathbf{e}_{s,y}^{(t)} = \mathbf{0}$.
\end{itemize}
The zero-sum noise generation is performed according to the CAPE protocol~\cite{imtiaz2021correlated}, and the details of the noise variance calculations are shown in Appendix \ref{sec:appenA}. Note that, the server receives \( S \) noisy vectors for each \( t \) and class \( k \), and performs the following aggregation: $\bar{\mathbf{x}}^{(k)}_t = \frac{1}{S} \sum_{s=1}^S \tilde{\mathbf{x}}^{(s,k)}_t$, $\bar{\mathbf{y}}^{\text{one-hot}(k)}_t = \frac{1}{S} \sum_{s=1}^S \tilde{\mathbf{y}}^{\text{one-hot}(s,k)}_t$. 
The label is decoded using $\bar{y}^{(k)}_t = \arg\max_i \bar{\mathbf{y}}^{\text{one-hot}(k)}_t[i]$. Finally, the synthetic data entries \( (\bar{\mathbf{x}}_t^{(k)}, \bar{y}_t^{(k)}) \) are collected over all \( k \in \{1, \dots, K\} \) and \( t \in \{1, \dots, T_k\} \) to form the final released dataset $\tilde{\mathbb{D}} = \left\{ (\bar{\mathbf{x}}_t, \bar{y}_t) \right\}_{t=1}^{T_s}$. The complete procedure is summarized in Algorithm~\ref{alg:decentralized-dpcda}.

\begin{algorithm}[t]
\caption{CAPE Assisted Federated DP-CDA: Privacy-Preserving Synthetic Data Generation}
\label{alg:decentralized-dpcda}

\KwIn{Local datasets $\{\mathbb{D}_s\}_{s=1}^S$ of size $N_s = N/S$, total synthetic samples per client $T_s = T$, mixture size $l$, number of classes $K$, clipping threshold $c$, noise variances $\tau_e^2$, $\tau_g^2$}
\KwOut{Aggregated synthetic dataset $\tilde{\mathbb{D}} = \{(\bar{x}_t, \bar{y}_t)\}_{t=1}^{T_s}$}


\ForPar{$s \gets 1$ \KwTo $S$}{
    Preprocess local dataset $\mathbb{D}_i$: apply z-score normalization and $\ell_2$ norm clipping\;
    
    \For{$t \gets 1$ \KwTo $T_s$}{
        For each class \( k \in \{1, \dots, K\} \), randomly select \( l \) samples from the local class-specific set:
        $\mathbf{X}_{s,k} = \{\mathbf{x}_{s,i} \in \mathbf{X}_s : y_{s,i} = k\}$
        
        Generate synthetic feature:\label{alg:decentralized-dpcda:step5}
        $
        \tilde{\mathbf{x}}^{(s,k)}_t = \frac{1}{l} \sum_{j=1}^{l} \mathbf{x}_{s,i_j} + \mathbf{e}_{s,x}^{(t)} + \mathbf{g}_{s,x}^{(t)}$\;
        
        Generate synthetic label (one-hot): 
        $\tilde{\mathbf{y}}^{\text{one-hot}(s,k)}_t = \frac{1}{l} \sum_{j=1}^{l} \mathbf{y}^{\text{one-hot}}_{s,i_j} + \mathbf{e}_{s,y}^{(t)} + \mathbf{g}_{s,y}^{(t)}$\;
        
        
        Store local synthetic sample $(\tilde{\mathbf{x}}^{(s,k)}_t, \tilde{\mathbf{y}}^{\text{one-hot}(s,k)}_t)$\;
    }
    
    Transmit $(\tilde{\mathbf{x}}^{(s,k)}_t, \tilde{\mathbf{y}}^{\text{one-hot}(s,k)}_t)$ to the central server\;
}

\For{$t \gets 1$ \KwTo $T_s$}{
    Aggregate features: $\bar{\mathbf{x}}^{(k)}_t = \frac{1}{S} \sum_{s=1}^S \tilde{\mathbf{x}}^{(s,k)}_t$\;
    
    Aggregate labels: $\bar{\mathbf{y}}^{\text{one-hot}(k)}_t = \frac{1}{S} \sum_{s=1}^S \tilde{\mathbf{y}}^{\text{one-hot}(s,k)}_t$\;
    
    Label decoding: $\bar{y}^{(k)}_t = \arg\max_i \bar{\mathbf{y}}^{\text{one-hot}(k)}_t[i]$\;
}

\Return Aggregated synthetic dataset: $\tilde{\mathbb{D}} = \left\{ (\bar{\mathbf{x}}_t, \bar{y}_t) \right\}_{t=1}^{T_s}$\;
\end{algorithm}

\begin{theorem}[Privacy of Algorithm~\ref{alg:decentralized-dpcda}]
\label{theorem:cape-dpcda}
Consider the CAPE-assisted Federated DP-CDA algorithm in the setting of Section~\ref{sec:Definitions}, where each client performs class-wise mixing of $l$ randomly-selected samples and adds noise according to the CAPE protocol for privacy (Step~\ref{alg:decentralized-dpcda:step5} of Algorithm~\ref{alg:decentralized-dpcda}). Then the released dataset $\tilde{\mathbb{D}} = \{ (\bar{\mathbf{x}}_t, \bar{y}_t) \}_{t=1}^{T}$ is $(\epsilon, \delta)$-differentially private for any $0 < \delta < 1$ and $\alpha \geq 3$, where $\epsilon = \min_{\alpha \in \{3, 4, \ldots\}} T\varepsilon'(\alpha) - \frac{\log\delta}{\alpha-1}$. Here $\varepsilon'(\alpha)= \frac{1}{\alpha - 1} \log (1 + p^2 {\alpha \choose 2} \min \left\{ 4(e^{\varepsilon(2)} - 1), 2 e^{\varepsilon(2)} \right\}$ + $4 G(\alpha) )$, $\varepsilon(\alpha) = \frac{\alpha}{l^2} \left( \frac{2c^2}{\tau_g^2} + \frac{1}{\tau_g^2} \right)$, and $G(\alpha)$ is given by $G(\alpha) = \sum_{j=3}^{\alpha} p^{j} {\alpha \choose j} \sqrt{B(2 \left\lfloor \frac{j}{2} \right\rfloor) \cdot B(\left\lceil \frac{j}{2} \right\rceil)}$ with $B(l) = \sum_{i=0}^{l} (-1)^{i} {l \choose i} e^{(i-1) \varepsilon(i)}$ and $p = \frac{lK}{N}$.
\end{theorem}

\begin{proof}
Let us consider the CAPE Assisted Federated DP-CDA mechanism, where each client $s \in \{1, 2, \dots, S\}$ generates synthetic samples by mixing $l$ class-specific data points as in Algorithm~\ref{alg:decentralized-dpcda}. Since $\sum_{s=1}^S \mathbf{e}_{s,x}^{(t)} = \mathbf{0}$ and $\sum_{s=1}^S \mathbf{e}_{s,y}^{(t)} = \mathbf{0}$ for each sample index $t$ due to CAPE protocol, the privacy analysis of the synthetic dataset generation reduces to analyzing the local noise component with variance $\tau_g^2$, similar to the centralized DP-CDA setup. Following Proposition~\ref{prop:rdp_gauss_mech}, the computation of the synthetic feature vector with added local noise is $(\alpha, \frac{\alpha}{2 (\frac{\tau_g}{\Delta_x})^2})$-RDP. Similarly, the synthetic label with added local noise is $(\alpha, \frac{\alpha}{2 (\frac{\tau_g}{\Delta_y})^2})$-RDP. By RDP composition (Proposition~\ref{prop:composition_rdp}), the total per-sample privacy cost is $\varepsilon(\alpha) = \frac{\alpha}{2} \left( \Delta_x^2 / \tau_g^2 + \Delta_y^2 / \tau_g^2 \right)$. Since $\|\mathbf{x}_i\|_2 \leq c$ and $\mathbf{y}_i$ are one-hot-encoded, we have $\Delta_x = \frac{2c}{l}$ and $\Delta_y = \frac{\sqrt{2}}{l}$. Substituting, we get $\varepsilon(\alpha) = \frac{\alpha}{l^2} \left( \frac{2c^2}{\tau_g^2} + \frac{1}{\tau_g^2} \right)$. Since the $l$ samples used for mixing are chosen uniformly at random from the local dataset, the overall mechanism follows the same sub-sampling amplification strategy as described in~\cite{wang2019subsampled}. Therefore, each synthetic sample generation follows $(\alpha, \varepsilon'(\alpha))$-RDP, where $\varepsilon'(\alpha)= \frac{1}{\alpha - 1} \log \left(1 + p^2 {\alpha \choose 2} \min \left\{ 4(e^{\varepsilon(2)} - 1), 2 e^{\varepsilon(2)} \right\} + 4 G(\alpha) \right)$. Additionally, $G(\alpha) = \sum_{j=3}^{\alpha} p^j {\alpha \choose j} \sqrt{B(2 \lfloor j/2 \rfloor) \cdot B(\lceil j/2 \rceil)}$, $B(l) = \sum_{i=0}^{l} (-1)^i {l \choose i} e^{(i-1) \varepsilon(i)}$, and $p = \frac{lK}{N}$ is the sampling probability for subsampling $l$ points from the dataset of size $N$. Since $T=T_s$ synthetic samples are generated in total, using the composition property of RDP and converting RDP to $(\epsilon, \delta)$-DP via Proposition~\ref{prop:rdp_dp}, the final privacy guarantee becomes $\epsilon = \min_{\alpha \in \{3, 4, \ldots\}} T \varepsilon'(\alpha) - \frac{\log\delta}{\alpha - 1}$.
\end{proof}

\begin{Rem}
    We note that if DP-CDA is employed in the conventional decentralized-data setting, i.e., without the CAPE protocol, each client must independently add noise to satisfy differential privacy. As a result, the standard deviation of the noise must be scaled by a factor of $\sqrt{S}$~\cite{imtiaz2021correlated} compared to the centralized scenario, where $S$ is the number of participating clients, to ensure an equivalent global privacy guarantee. This increased noise severely degrades the utility of the generated synthetic data, especially as the number of clients grows. We refer the reader to Section~\ref{sec:experiments} for an empirical validation and~\cite{imtiaz2021correlated} for theoretical details.
\end{Rem}
\begin{Rem}
    The computational complexity of the CAPE Assisted Federated DP-CDA algorithm is \( \mathcal{O}(T \cdot S \cdot l \cdot d_x) \). Please refer to the complexity calculation in Appendix \ref{sec:appenA}.
\end{Rem}

\section{Experimental Results}\label{sec:experiments}

\begin{table*}[t]
\centering
\caption{Test accuracy (\%) under different privacy budgets and order of mixture (\(l\)) for CAPE-assisted Federated DP-CDA and Conventional Federated DP-CDA on MNIST and FashionMNIST datasets with \(S = 10\) clients.}
\label{tab:fed-results-combined}
\resizebox{\textwidth}{!}{
\begin{tabular}{c c}
\hline
\hline
\textbf{(a) CAPE-assisted Federated DP-CDA} & \textbf{(b) Conventional Federated DP-CDA} \\
\begin{tabular}{c | ccc | ccc}
\hline
\multirow{2}{*}{\textbf{\(l\)}} & \multicolumn{3}{c}{\textbf{MNIST}} & \multicolumn{3}{c}{\textbf{FashionMNIST}} \\
 & \(\varepsilon = \infty\) & \(\varepsilon = 20\) & \(\varepsilon = 10\) & \(\varepsilon = \infty\) & \(\varepsilon = 20\) & \(\varepsilon = 10\) \\
\hline
1   & 98.85 & 10.80 & 10.81 & 90.64 & 38.86 & 36.19 \\
2   & 97.46 & 67.22 & 70.76 & 88.51 & 67.12 & 66.55 \\
4   & 94.62 & 78.08 & 78.07 & 84.54 & 67.91 & 67.48 \\
8   & 90.45 & 78.02 & 76.73 & 78.81 & 67.19 & 67.03 \\
16  & 86.07 & 76.91 & 75.85 & 73.70 & 66.52 & 66.53 \\
32  & 82.37 & 75.77 & 67.38 & 69.70 & 65.42 & 66.29 \\
64  & 79.62 & 71.75 & 63.19 & 65.79 & 65.38 & 64.98 \\
128 & 78.29 & 66.56 & 54.31 & 64.35 & 65.32 & 63.99 \\
256 & 76.93 & 36.20 & 34.03 & 62.73 & 59.66 & 62.03 \\
512 & 76.03 & 25.97 & 15.75 & 61.81 & 52.43 & 45.74 \\
\hline
\end{tabular}
&
\begin{tabular}{c | ccc | ccc}
\hline
\multirow{2}{*}{\textbf{\(l\)}} & \multicolumn{3}{c}{\textbf{MNIST}} & \multicolumn{3}{c}{\textbf{FashionMNIST}} \\
 & \(\varepsilon = \infty\) & \(\varepsilon = 20\) & \(\varepsilon = 10\) & \(\varepsilon = \infty\) & \(\varepsilon = 20\) & \(\varepsilon = 10\) \\
\hline
1   & 86.80 & 22.45 & 23.05 & 79.48 & 32.89 & 27.82 \\
2   & 83.83 & 61.04 & 59.40 & 75.72 & 61.09 & 58.68 \\
4   & 81.09 & 67.49 & 66.19 & 71.76 & 63.30 & 61.01 \\
8   & 79.70 & 65.84 & 65.01 & 70.55 & 61.62 & 61.44 \\
16  & 78.92 & 61.87 & 63.49 & 69.40 & 60.98 & 60.21 \\
32  & 78.08 & 59.92 & 58.71 & 68.03 & 59.32 & 57.62 \\
64  & 77.24 & 56.00 & 54.25 & 68.00 & 56.55 & 56.95 \\
128 & 78.76 & 50.12 & 45.92 & 68.08 & 54.31 & 50.60 \\
256 & 77.06 & 40.43 & 36.08 & 66.88 & 44.15 & 46.40 \\
512 & 77.91 & 27.58 & 30.29 & 67.19 & 41.28 & 34.90 \\
\hline
\end{tabular}
\\
\end{tabular}
}
\end{table*}

We evaluate the effectiveness of our proposed CAPE-assisted Federated DP-CDA method on two benchmark datasets: MNIST~\cite{lecun2002gradient} and FashionMNIST~\cite{xiao2017fashion}. In each experiment, the data is partitioned across \( S = 10 \) clients, and synthetic data is generated locally at each client using a range of $l$ values: \( l \in \{1, 2, 4, \dots, 512\} \). Experiments are conducted under three levels of differential privacy budgets: \( \varepsilon \in \{\infty, 20, 10\} \). For model training and evaluating the utility of the synthetic data generated through our proposed framework, we used a small convolutional neural network (CNN) and measured its test accuracy on the test partition of the corresponding real dataset. The details about the CNN architecture are provided in Appendix \ref{sec:appenA}.

\noindent\textbf{Performance on Test Partitions of Real Datasets in CAPE Assisted Federated DP-CDA. }Note that the model is trained on the synthetic data generated using the proposed CAPE Assisted Federated DP-CDA scheme, and then tested on the test partition of the corresponding real datasets. The results obtained from the proposed algorithm are summarized in Table~\ref{tab:fed-results-combined}(a), which shows the test accuracy of models trained on the aggregated synthetic datasets. Recall that the correlated noise is canceled out during the final data aggregation at the central aggregator for our proposed scheme. We intentionally set the local noise to ensure that the noise variance of the $\bar{\mathbf{x}}^{(k)}_t$ and $\bar{\mathbf{y}}^{\text{one-hot}(k)}_t$ match that of the pooled-data or centralized scenarios. As a result, the same test accuracy (\%) can be achieved as in the centralized DP-CDA~\cite{saha2024dp} through our proposed CAPE Assisted Federated DP-CDA. We refer the reader to~\cite{imtiaz2021correlated} for a formal analysis of this.

\noindent\textbf{Performance on Test Partitions of Real Datasets in Conventional Federated DP-CDA. }As previously mentioned, in conventional Federated DP-CDA, to achieve the same level of privacy as in the centralized scenario, we need to add noise with a standard deviation scaled by a factor of \(\sqrt{S}\) compared to the centralized case~\cite{imtiaz2021correlated}. As a result, the synthetic dataset generated using conventional Federated DP-CDA is much noisier compared to the proposed CAPE Assisted Federated DP-CDA. Consequently, it is expected that the utility of the dataset in the conventional case will be lower -- the results are compared in Table \ref{tab:fed-results-combined}(b).

\noindent\textbf{Performance Comparison Between Conventional and CAPE Assisted Federated DP-CDA. } 
To assess the advantages of our proposed CAPE Assisted Federated DP-CDA algorithm, we compare its performance with the conventional Federated DP-CDA approach (i.e., no correlated noise) across two datasets—MNIST and FashionMNIST—under varying privacy budgets (\( \epsilon = \infty \), 20, and 10). As mentioned before, the results are presented in Table~\ref{tab:fed-results-combined}(a) (proposed scheme) and Table~\ref{tab:fed-results-combined}(b) (conventional scheme). Considering Table~\ref{tab:fed-results-combined}, it is evident that the performance of the proposed CAPE Assisted Federated DP-CDA is much better than the conventional Federated DP-CDA.
Specifically, for the MNIST dataset, at a strict privacy budget of \( \epsilon = 10 \), the proposed scheme achieves up to 78.07\% test accuracy for \( l = 4 \), while the conventional method only reaches 66.19\%, showing an improvement of nearly 12\%. Even for moderate privacy (\( \epsilon = 20 \)), the proposed approach achieves 78.08\%, compared to 67.49\% for the conventional method. A similar trend is observed for FashionMNIST. At \( \epsilon = 10 \), the proposed scheme maintains a strong performance of 67.48\%, while the conventional method drops to 61.01\% for \( l = 4 \). At \( \epsilon = 20 \), the proposed method continues to outperform with 67.91\% versus 63.30\%, showing that the proposed method consistently retains better generalization ability under both moderate and strict privacy budgets.

\begin{figure}[t]
    \centering
    \includegraphics[width=1.0\columnwidth]{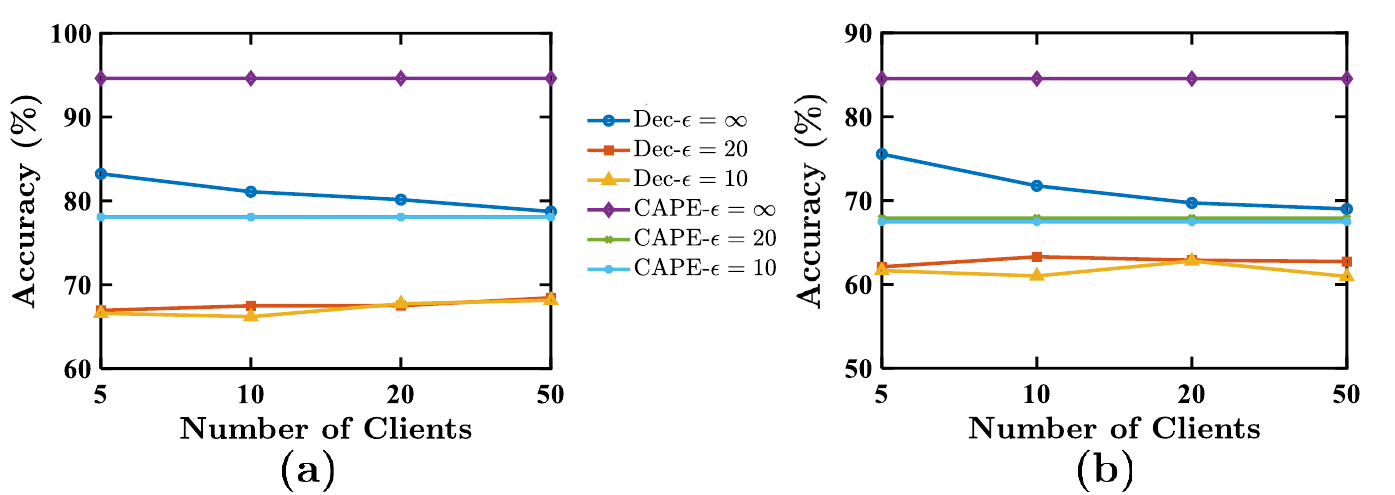}
    \caption{Effect of the number of participating clients on dataset utility. Comparison of conventional decentralized DP-CDA and CAPE-assisted federated DP-CDA on (a) MNIST and (b) FashionMNIST datasets.}
    \label{fig:fig2}
    \vspace{-0.5cm}
\end{figure}

\noindent\textbf{Impact of Number of Clients on Model Utility. }
To understand the impact of the number of participating clients on synthetic data generation and model performance, we analyze how the utility varies as $S$ increases. In this experiment, we fix the order of mixture to \( l = 4 \) and evaluate model performance under three privacy budgets: \( \varepsilon \in \{\infty, 20, 10\} \), across varying numbers of clients \( S \in \{5, 10, 20, 50\} \). The results are shown in Figure~\ref{fig:fig2}, where each subplot shows the accuracy trend for a specific dataset (MNIST and FashionMNIST), comparing the conventional federated DP-CDA and the proposed CAPE Assisted Federated DP-CDA.

Figure~\ref{fig:fig2} shows how the number of clients affects model utility across different privacy levels. We first observe that the CAPE Assisted Federated DP-CDA method demonstrates consistent and robust performance across all participating clients. This is because CAPE leverages correlated noise, which allows each client to preserve local privacy while ensuring that the aggregated noise cancels out at the server. As a result, the utility of the final model remains unaffected by the number of participating clients. For example, on the MNIST dataset, accuracy under CAPE remains stable at 94.62\% for the non-private case ($\varepsilon = \infty$), and approximately 78.08\% for both $\varepsilon = 20$ and $\varepsilon = 10$, regardless of whether there are 5, 10, 20, or 50 clients. A similar trend is observed for the FashionMNIST dataset. 

In contrast, the conventional decentralized DP-CDA approach (without CAPE) shows noticeable degradation in model utility as the number of clients increases. This trend is more evident in the non-private setting, where accuracy drops due to reduced local data per client. For instance, in the MNIST dataset, accuracy decreases from 83.23\% to 78.73\% as the number of clients increases from 5 to 50 under $\varepsilon = \infty$. With privacy constraints, although the degradation is less severe, there is still a small decline. A similar pattern is observed for FashionMNIST. 
These results highlight a key advantage of the proposed approach: by injecting correlated noise, CAPE avoids the utility loss typically associated with scaling federated systems while maintaining strong privacy.

\begin{table}[t]
\centering
\caption{Performance comparison of the proposed algorithm with existing approaches in terms of privacy and utility (accuracy).}
\label{tab:2}
\resizebox{0.6\columnwidth}{!}{ 
\begin{tabular}{ccc cc}
\hline
\hline
\multirow{2}{*}{Algorithm} & \multicolumn{2}{c}{\textbf{MNIST}} & \multicolumn{2}{c}{\textbf{FashionMNIST}} \\ \cline{2-5}
 & $\epsilon$ = 20 & $\epsilon$ = 10 & $\epsilon$ = 20 & $\epsilon$ = 10 \\ \hline
Random Projection~\cite{xu2017dppro}          & 0.104 & 0.100 & -     & -     \\ 
Local Perturbation~\cite{agrawal2000privacy}   & 0.098 & 0.098 & -     & -     \\ 
DP-Mix~\cite{lee2019synthesizing}              & 0.800 & 0.782 & -     & -     \\ 
DP-CDA~\cite{saha2024dp}                       & 0.796 & 0.795 & 0.685 & 0.680 \\ 
Federated DP-CDA (conventional)                & 0.675 & 0.662 & 0.633 & 0.610 \\
CAPE Assisted Federated DP-CDA                 & 0.796 & 0.795 & 0.685 & 0.680 \\
\hline
\end{tabular}
}
\end{table}

\noindent\textbf{Comparison with Existing Approaches. }
To assess the effectiveness of our proposed framework, we compare the performance of \textit{Federated DP-CDA} and its CAPE-enhanced variant against several existing privacy-preserving data publishing methods. Table~\ref{tab:2} summarizes the results in terms of utility (test accuracy) and different privacy budgets (\( \epsilon \)) across two benchmark datasets: MNIST and FashionMNIST. Classical local privacy mechanisms such as Random Projection~\cite{xu2017dppro} and Local Perturbation~\cite{agrawal2000privacy} yield very poor accuracy, typically below 10\% across all datasets and privacy levels. These methods ensure strong privacy but at the expense of extreme utility loss, making them unsuitable for practical machine learning tasks. DP-Mix~\cite{lee2019synthesizing}, a notable synthetic data generation method, offers a better trade-off. It achieves an accuracy of 0.800 and 0.782 on MNIST for \( \epsilon = 20 \) and \( \epsilon = 10 \), respectively. 
The recently proposed centralized DP-CDA~\cite{saha2024dp} performs competitively with DP-Mix, achieving similar or slightly better accuracy on all datasets. Notably, it provides strong utility even at tight privacy budgets (\( \epsilon = 10 \)) -- achieving 0.795 on MNIST and 0.680 on FashionMNIST. When extended to a federated setting, Federated DP-CDA experiences some performance degradation due to increased local noise (necessitated by privacy preservation across distributed clients). For example, on MNIST with \( \epsilon = 20 \), accuracy drops from 0.796 (centralized) to 0.675 (federated). Similar declines are observed on FashionMNIST, demonstrating the inherent privacy-utility trade-off in decentralized scenarios.

To address this, we integrate CAPE into our federated framework. As such, the proposed CAPE Assisted Federated DP-CDA restores accuracy levels to match those of the centralized DP-CDA, achieving 0.795 on MNIST and 0.680 on FashionMnist at \( \epsilon = 10 \). This confirms that CAPE effectively mitigates the utility degradation caused by local noise in federated settings, without requiring additional computational overhead.
Overall, the results validate that the proposed CAPE Assisted Federated DP-CDA achieves centralized-level utility, formal differential privacy guarantees, and efficient runtime, outperforming the existing methods.

\section{Conclusion}
In this work, we extended the Differentially Private Class-Centric Data Aggregation (DP-CDA) algorithm to federated settings and addressed the utility degradation from local noise injection by integrating the CAPE protocol. This decomposition of local noise into independent and jointly generated components enables each client to meet differential privacy requirements, while allowing the aggregator to recover high-utility synthetic data with reduced global noise. Our theoretical and empirical results on MNIST and FashionMNIST show that CAPE-assisted Federated DP-CDA consistently outperforms conventional decentralized mechanisms and achieves utility comparable to centralized DP-CDA without additional computational cost. These findings demonstrate the potential of our approach for scalable and trustworthy privacy-preserving synthetic data generation in federated environments, with future work targeting more complex data modalities and heterogeneous (non-IID) data distributions.

\bibliographystyle{ieeetr}
\bibliography{references}

\begin{thebibliography}{10}

\bibitem{shokri2017membership}
R.~Shokri, M.~Stronati, C.~Song, and V.~Shmatikov, ``Membership inference attacks against machine learning models,'' in {\em 2017 IEEE symposium on security and privacy (SP)}, pp.~3--18, IEEE, 2017.

\bibitem{nasr2019comprehensive}
M.~Nasr, R.~Shokri, and A.~Houmansadr, ``Comprehensive privacy analysis of deep learning: Passive and active white-box inference attacks against centralized and federated learning,'' in {\em 2019 IEEE symposium on security and privacy (SP)}, pp.~739--753, IEEE, 2019.

\bibitem{carlini2019secret}
N.~Carlini, C.~Liu, {\'U}.~Erlingsson, J.~Kos, and D.~Song, ``The secret sharer: Evaluating and testing unintended memorization in neural networks,'' in {\em 28th USENIX security symposium (USENIX security 19)}, pp.~267--284, 2019.

\bibitem{ganju2018property}
K.~Ganju, Q.~Wang, W.~Yang, C.~A. Gunter, and N.~Borisov, ``Property inference attacks on fully connected neural networks using permutation invariant representations,'' in {\em Proceedings of the 2018 ACM SIGSAC conference on computer and communications security}, pp.~619--633, 2018.

\bibitem{chen2023federated}
H.~Chen and H.~Vikalo, ``Federated learning in non-iid settings aided by differentially private synthetic data,'' in {\em Proceedings of the IEEE/CVF Conference on Computer Vision and Pattern Recognition}, pp.~5027--5036, 2023.

\bibitem{augenstein2019generative}
S.~Augenstein, H.~B. McMahan, D.~Ramage, S.~Ramaswamy, P.~Kairouz, M.~Chen, R.~Mathews, {\em et~al.}, ``Generative models for effective ml on private, decentralized datasets,'' {\em 2020 International Conference on Learning Representations (ICLR)}, 2020.

\bibitem{li2020federated}
T.~Li, A.~K. Sahu, A.~Talwalkar, and V.~Smith, ``Federated learning: Challenges, methods, and future directions,'' {\em IEEE signal processing magazine}, vol.~37, no.~3, pp.~50--60, 2020.

\bibitem{yang2019federated}
Q.~Yang, Y.~Liu, T.~Chen, and Y.~Tong, ``Federated machine learning: Concept and applications,'' {\em ACM Transactions on Intelligent Systems and Technology (TIST)}, vol.~10, no.~2, pp.~1--19, 2019.

\bibitem{mcmahan2017communication}
B.~McMahan, E.~Moore, D.~Ramage, S.~Hampson, and B.~A. y~Arcas, ``Communication-efficient learning of deep networks from decentralized data,'' in {\em Artificial intelligence and statistics}, pp.~1273--1282, PMLR, 2017.

\bibitem{konevcny2016federated}
J.~Kone{\v{c}}n{\`y}, H.~B. McMahan, F.~X. Yu, P.~Richt{\'a}rik, A.~T. Suresh, and D.~Bacon, ``Federated learning: Strategies for improving communication efficiency,'' {\em arXiv preprint arXiv:1610.05492}, 2016.

\bibitem{geiping2020inverting}
J.~Geiping, H.~Bauermeister, H.~Dr{\"o}ge, and M.~Moeller, ``Inverting gradients-how easy is it to break privacy in federated learning?,'' {\em Advances in neural information processing systems}, vol.~33, pp.~16937--16947, 2020.

\bibitem{aouedi2022handling}
O.~Aouedi, A.~Sacco, K.~Piamrat, and G.~Marchetto, ``Handling privacy-sensitive medical data with federated learning: challenges and future directions,'' {\em IEEE journal of biomedical and health informatics}, vol.~27, no.~2, pp.~790--803, 2022.

\bibitem{jin2021cafe}
X.~Jin, P.-Y. Chen, C.-Y. Hsu, C.-M. Yu, and T.~Chen, ``Cafe: Catastrophic data leakage in vertical federated learning,'' {\em Advances in neural information processing systems}, vol.~34, pp.~994--1006, 2021.

\bibitem{xu2024tapfed}
R.~Xu, B.~Li, C.~Li, J.~B. Joshi, S.~Ma, and J.~Li, ``Tapfed: Threshold secure aggregation for privacy-preserving federated learning,'' {\em IEEE Transactions on Dependable and Secure Computing}, vol.~21, no.~5, pp.~4309--4323, 2024.

\bibitem{chang2023privacy}
Y.~Chang, K.~Zhang, J.~Gong, and H.~Qian, ``Privacy-preserving federated learning via functional encryption, revisited,'' {\em IEEE Transactions on Information Forensics and Security}, vol.~18, pp.~1855--1869, 2023.

\bibitem{asoodeh2021differentially}
S.~Asoodeh, W.-N. Chen, F.~P. Calmon, and A.~{\"O}zg{\"u}r, ``Differentially private federated learning: An information-theoretic perspective,'' in {\em 2021 IEEE international symposium on information theory (ISIT)}, pp.~344--349, IEEE, 2021.

\bibitem{xu2021fedv}
R.~Xu, N.~Baracaldo, Y.~Zhou, A.~Anwar, J.~Joshi, and H.~Ludwig, ``Fedv: Privacy-preserving federated learning over vertically partitioned data,'' in {\em Proceedings of the 14th ACM workshop on artificial intelligence and security}, pp.~181--192, 2021.

\bibitem{geyer2017differentially}
R.~C. Geyer, T.~Klein, and M.~Nabi, ``Differentially private federated learning: A client level perspective,'' {\em arXiv preprint arXiv:1712.07557}, 2017.

\bibitem{saha2024dp}
U.~Saha, T.~M. Tonoy, and H.~Imtiaz, ``Dp-cda: An algorithm for enhanced privacy preservation in dataset synthesis through randomized mixing,'' {\em arXiv preprint arXiv:2411.16121}, 2024.

\bibitem{imtiaz2021correlated}
H.~Imtiaz, J.~Mohammadi, R.~Silva, B.~Baker, S.~M. Plis, A.~D. Sarwate, and V.~D. Calhoun, ``A correlated noise-assisted decentralized differentially private estimation protocol, and its application to fmri source separation,'' {\em IEEE Transactions on Signal Processing}, vol.~69, pp.~6355--6370, 2021.

\bibitem{saha2024privacy}
S.~Saha and H.~Imtiaz, ``Privacy-preserving non-negative matrix factorization with outliers,'' {\em ACM Transactions on Knowledge Discovery from Data}, vol.~18, no.~3, pp.~1--26, 2024.

\bibitem{tasnim2023approximating}
N.~Tasnim, J.~Mohammadi, A.~D. Sarwate, and H.~Imtiaz, ``Approximating functions with approximate privacy for applications in signal estimation and learning,'' {\em Entropy}, vol.~25, no.~5, p.~825, 2023.

\bibitem{wang2019subsampled}
Y.-X. Wang, B.~Balle, and S.~P. Kasiviswanathan, ``Subsampled r{\'e}nyi differential privacy and analytical moments accountant,'' in {\em The 22nd international conference on artificial intelligence and statistics}, pp.~1226--1235, PMLR, 2019.

\bibitem{lecun2002gradient}
Y.~LeCun, L.~Bottou, Y.~Bengio, and P.~Haffner, ``Gradient-based learning applied to document recognition,'' {\em Proceedings of the IEEE}, vol.~86, no.~11, pp.~2278--2324, 2002.

\bibitem{xiao2017fashion}
H.~Xiao, K.~Rasul, and R.~Vollgraf, ``Fashion-mnist: a novel image dataset for benchmarking machine learning algorithms,'' {\em arXiv preprint arXiv:1708.07747}, 2017.

\bibitem{xu2017dppro}
C.~Xu, J.~Ren, Y.~Zhang, Z.~Qin, and K.~Ren, ``Dppro: Differentially private high-dimensional data release via random projection,'' {\em IEEE Transactions on Information Forensics and Security}, vol.~12, no.~12, pp.~3081--3093, 2017.

\bibitem{agrawal2000privacy}
R.~Agrawal and R.~Srikant, ``Privacy-preserving data mining,'' in {\em Proceedings of the 2000 ACM SIGMOD international conference on Management of data}, pp.~439--450, 2000.

\bibitem{lee2019synthesizing}
K.~Lee, H.~Kim, K.~Lee, C.~Suh, and K.~Ramchandran, ``Synthesizing differentially private datasets using random mixing,'' in {\em 2019 IEEE International Symposium on Information Theory (ISIT)}, pp.~542--546, IEEE, 2019.

\bibitem{dwork2006calibrating}
C.~Dwork, F.~McSherry, K.~Nissim, and A.~Smith, ``Calibrating noise to sensitivity in private data analysis,'' in {\em Theory of Cryptography: Third Theory of Cryptography Conference, TCC 2006, New York, NY, USA, March 4-7, 2006. Proceedings 3}, pp.~265--284, Springer, 2006.

\bibitem{dwork2014algorithmic}
C.~Dwork, A.~Roth, {\em et~al.}, ``The algorithmic foundations of differential privacy,'' {\em Foundations and Trends{\textregistered} in Theoretical Computer Science}, vol.~9, no.~3--4, pp.~211--407, 2014.

\bibitem{mcsherry2007mechanism}
F.~McSherry and K.~Talwar, ``Mechanism design via differential privacy,'' in {\em 48th Annual IEEE Symposium on Foundations of Computer Science (FOCS'07)}, pp.~94--103, IEEE, 2007.

\bibitem{mironov2017renyi}
I.~Mironov, ``R{\'e}nyi differential privacy,'' in {\em 2017 IEEE 30th computer security foundations symposium (CSF)}, pp.~263--275, IEEE, 2017.

\end{thebibliography}
\newpage
\clearpage

\appendix
\section{Relevant Definitions, Theorems, and Proofs}\label{sec:appenA}
\appendix
In this section, we present some definitions and theorems relevant to our proposed method. Additionally, we show the detailed calculations of the noise variance for the two noise terms, computational complexity calculation, and details of the tiny CNN used for model training.

\begin{definition}[($\epsilon,\delta$)-DP \cite{dwork2006calibrating}]
  An algorithm $f : \mathcal{D} \mapsto \mathcal{T}$ provides ($\epsilon,\delta$)-differential privacy (($\epsilon,\delta$)-DP) if $\Pr(f(\mathbb{D})\in \mathcal{S}) \le \delta + e^\epsilon \Pr(f(\mathbb{D}^\prime)\in \mathcal{S} )$ for all measurable $\mathcal{S} \subseteq \mathcal{T} $ and for all neighboring datasets $\mathbb{D},\mathbb{D}^\prime \in \mathcal{D}$.
\end{definition}

Here, $\epsilon > 0,\ 0 < \delta < 1$ are the privacy parameters, and determine how the algorithm will perform in providing privacy/utility. The parameter $\epsilon$ indicates how much the algorithm's output deviates in probability when we replace one single person's data with another. The parameter $\delta$ indicates the probability that the privacy mechanism fails to give the guarantee of $\epsilon$. Intuitively, higher privacy results in poor utility. That is, smaller $\epsilon$ and $\delta$ guarantee more privacy, but lower utility. There are several mechanisms to implement DP: Gaussian \cite{dwork2006calibrating}, Laplace mechanism \cite{dwork2014algorithmic}, random sampling, and exponential mechanism~\cite{mcsherry2007mechanism} are well-known. Among the additive noise mechanisms, the noise standard deviation is scaled by the privacy budget and the sensitivity of the function.\\

\begin{definition}[$\ell_2$ sensitivity \cite{dwork2006calibrating}] The $\ell_2$- sensitivity of vector valued function $f(\mathbb{D})$ is $\Delta := \max_{\mathbb{D},\ \mathbb{D}^\prime} \Vert f(\mathbb{D}) - f(\mathbb{D}') \Vert_2$, where $\mathbb{D}$ and $\mathbb{D}^\prime$ are neighboring datasets.
\end{definition} 
The $\ell_2$ sensitivity of a function gives the upper bound of how much the function can change if one sample at the input is changed. Consequently, it dictates the amount of randomness/perturbation needed at the function's output to guarantee DP. In other words, it captures the maximum change in the output by changing any one user in the worst-case scenario.\\

\begin{definition}[Gaussian Mechanism \cite{dwork2014algorithmic}] Let $f: \mathcal{D} \mapsto \mathcal{R}^D$ be an arbitrary function with $\ell_{2}$ sensitivity $\Delta$. The Gaussian mechanism with parameter $\tau$ adds noise from $\mathcal{N}(0,\tau^2)$ to each of the $D$ entries of the output and satisfies $(\epsilon,\delta)$-DP for $\epsilon \in (0,1)$ and $\delta \in (0,1)$, if $\tau \geq \frac{\Delta}{\epsilon} \sqrt{2\log\frac{1.25}{\delta}}$.
\end{definition} 
Here, $(\epsilon,\delta)$-DP is guaranteed by adding noise drawn form $\mathcal{N}(0,\tau^2)$ distribution. Note that, there are an infinite number of combinations of $(\epsilon,\delta)$ for a given $\tau^2$~\cite{imtiaz2021correlated}.\\

\begin{definition}[R\'enyi Differential Privacy (RDP) \cite{mironov2017renyi}] A randomized algorithm $f$ : $\mathcal{D} \mapsto \mathcal{T}$ is $(\alpha,\epsilon_{r})$-R\'enyi differentially private if, for any adjacent $\mathbb{D},\ \mathbb{D}^\prime \in \mathcal{D}$, the following holds: $D_{\alpha}(\mathcal{A}(\mathbb{D})\ || \mathcal{A}(\mathbb{D}^\prime)) \leq \epsilon_{r}$. Here, $D_{\alpha}(P(x)||Q(x))=\frac{1}{\alpha-1}\log\mathbb{E}_{x \sim Q} \Big(\frac{P(x)}{Q(x)}\Big)^{\alpha}$ and $P(x)$ and $Q(x)$ are probability density functions defined on $\mathcal{T}$.
\end{definition} 

\begin{prop}[From RDP to DP~\cite{mironov2017renyi}]\label{prop:rdp_dp}
If $f$ is an $(\alpha,\epsilon_r)$-RDP mechanism, it also satisfies $(\epsilon_r+\frac{\log 1/\delta}{\alpha-1},\delta)$-DP for any $0<\delta<1$.
\end{prop}

\begin{prop}[Composition of RDP~\cite{mironov2017renyi}]\label{prop:composition_rdp}
Let $f_1:\mathcal{D}\rightarrow \mathcal{R}_1$ be $(\alpha,\epsilon_1)$-RDP and $f_2:{\mathcal{R}_1} \times \mathcal{D} \rightarrow \mathcal{R}_2$ be $(\alpha,\epsilon_2)$-RDP, then the mechanism defined as $(X_1,X_2)$, where $X_1 \sim f_1(\mathbb{D})$ and $X_2 \sim f_2(X_1,\mathbb{D})$ satisfies $(\alpha,\epsilon_1 + \epsilon_2)$-RDP.
\end{prop}

\begin{prop}[RDP and Gaussian Mechanism~\cite{mironov2017renyi}]\label{prop:rdp_gauss_mech}
If $f$ has $\ell_2$ sensitivity 1, then the Gaussian mechanism $\mathcal{G}_{\sigma}f(\mathbb{D})=f(\mathbb{D})+e$ where $e \sim \mathcal{N}(0,\sigma^2)$ satisfies $(\alpha,\frac{\alpha}{2\sigma^2})$-RDP. Also, a composition of $T$ such Gaussian mechanisms satisfies $(\alpha,\frac{\alpha T}{2 \sigma^2})$-RDP.
\end{prop}

\noindent\textbf{Variance Calculation of Correlation-Assisted Private Estimation (CAPE) in Federated DP-CDA~\cite{imtiaz2021correlated}. }In federated settings, preserving differential privacy often requires injecting large noise locally at each client before sending any output to the server. For example, to compute an \((\varepsilon, \delta)\)-differentially private estimate of a function \(f(\mathbf{x})\), the standard approach is to apply the Gaussian mechanism, where the noise scale is proportional to the sensitivity of the function.

In a \textit{centralized setting}, estimating the mean function \(f(\mathbf{x}) = \frac{1}{N} \sum_{i=1}^N x_i\), where \(x_i \in [0,1]\), requires adding noise \(\varepsilon_{\text{pool}} \sim \mathcal{N}(0, \tau_{\text{pool}}^2)\) with standard deviation:
\[
\tau_{\text{pool}} = \frac{1}{N} \sqrt{2 \log \frac{1.25}{\delta}}.
\]

However, in \textit{decentralized (federated) settings}, where the data is split across \(S\) clients and each client holds \(N_s = N/S\) samples, this sensitivity is now relative to smaller local datasets. To achieve the same level of privacy locally, each client must add noise with a standard deviation:
\[
\tau_s = \frac{1}{N_s} \sqrt{2 \log \frac{1.25}{\delta}} = \sqrt{S} \cdot \tau_{\text{pool}}.
\]
This scaling of the noise by \(\sqrt{S}\) causes a significant degradation in utility when the noisy local estimates are averaged at the server. This is because the variance of the aggregate estimator becomes:
\[
\text{Var}(\hat{f}) = \frac{1}{S^2} \sum_{s=1}^S \tau_s^2 = \frac{1}{S^2} \cdot S \cdot \tau_s^2 = \frac{\tau_s^2}{S} = \tau_{\text{pool}}^2 \cdot S,
         \]
which is \(S\)-times worse than the variance in the centralized case. To overcome this limitation, we employ the \textit{Correlation-Assisted Private Estimation (CAPE)} protocol. The key idea behind CAPE is to split the noise at each client into \textbf{two components}:
\begin{itemize}
    \item A \textbf{local noise} term \( g_s \sim \mathcal{N}(0, \tau_g^2) \), generated independently at each client.
    \item A \textbf{correlated noise} term \( e_s \sim \mathcal{N}(0, \tau_e^2) \), jointly generated across all clients such that $\sum_{s=1}^S e_s = 0$.
\end{itemize}
Each client \(s\) then transmits a privatized estimate of the form: $\hat{a}_s = f(\mathbf{x}_s) + e_s + g_s$ to the aggregator. Due to the zero-sum constraint on the correlated noise terms, they cancel out during aggregation. The aggregator computes:
\begin{align*}
\hat{a}_{\text{cape}}   &= \frac{1}{S} \sum_{s=1}^S \hat{a}_s \\
                        &= \frac{1}{S} \sum_{s=1}^S \left( f(\mathbf{x}_s) + e_s + g_s \right) \\
                        &= f(\mathbf{x}) + \frac{1}{S} \sum_{s=1}^S g_s,    
\end{align*}
as \( \sum_{s=1}^S e_s = 0 \). Therefore, the final noise at the server only stems from the local \( g_s \) terms.

If each local noise is calibrated such that \( \tau_g^2 = \tau_{\text{pool}}^2 \), then the variance of the CAPE estimator becomes:
\[
\tau_{\text{cape}}^2 = \frac{1}{S^2} \sum_{s=1}^S \tau_g^2 = \frac{S \cdot \tau_{\text{pool}}^2}{S^2} = \tau_{\text{pool}}^2,
\]
which is identical to the variance in the centralized setting. Thus, CAPE achieves the same privacy guarantee at the client level while preserving the centralized utility at the server level. This decomposition allows each site to satisfy local \((\varepsilon, \delta)\)-DP through the combined noise \( e_s + g_s \), while leveraging the correlated noise structure to cancel out part of the privacy cost during aggregation. Consequently, CAPE effectively removes the \(\sqrt{S}\)-factor penalty present in conventional decentralized DP and restores the utility of the global model to match the centralized baseline.

\noindent\textbf{Computational Complexity of the proposed algorithm. }To analyze the computational complexity of the proposed CAPE Assisted Federated DP-CDA algorithm, we consider the operations performed at both the client side and aggregator side. Each of the \( S \) clients contributes to the generation of \( T_s \) synthetic samples, where each sample is synthesized by randomly selecting \( l \) samples of dimension \( d_x \) from the same class. At the client side, for each synthetic sample, computing the average of \( l \) feature vectors and adding two noise components (one local and one correlated) requires \( \mathcal{O}(l d_x) \) operations. Similarly, for one-hot encoded labels of dimension \( d_y \), averaging and adding noise requires an additional \( \mathcal{O}(l d_y) \) operations. In most practical scenarios, \( d_x \gg d_y\). Thus, the per-sample computation cost per client is \( \mathcal{O}(ld_x) \), and across all \( T = T_s \) synthetic samples and all \( S \) clients, the total client-side complexity becomes \( \mathcal{O}(T \cdot S \cdot l \cdot d_x) \).

At the server side, for each of the \( T \) synthetic samples, aggregation involves computing the average of \( S \) feature vectors and label vectors, followed by decoding the final label using an \(\arg\max\) operation. This results in a per-sample aggregation cost of \( \mathcal{O}(Sd_x) \), and a total server-side cost of \( \mathcal{O}(T \cdot S \cdot d_x) \), which is of the same order as the client-side cost. Therefore, the overall computational complexity of the CAPE Assisted Federated DP-CDA algorithm is \( \mathcal{O}(T \cdot S \cdot l \cdot d_x) \), which scales linearly with the number of synthetic samples, the number of clients, and the degree of mixture.

\noindent\textbf{Model Architecture. }The CNN model consists of a single convolutional block followed by a sequence of fully connected layers. The convolutional block includes a 2D convolutional layer with 32 filters of size $5 \times 5$, stride 1, and padding 2, followed by a ReLU activation, batch normalization, and a $2 \times 2$ max pooling layer with stride 2. The resulting feature map is flattened and passed through two fully connected layers, each with 100 hidden units, ReLU activations, and dropout with a rate of 0.5. The final output layer is a fully connected layer with 10 output units corresponding to the number of classes. We note that the model was trained using the Adam optimizer with a fixed learning rate of \(1 \times 10^{-5}\). Cross-entropy loss was used as the objective function for training. The network was trained for a total of 100 epochs. All training runs were performed on a single GPU using the PyTorch framework. \\




\end{document}